\documentclass{article}
\usepackage{float}
\usepackage{xcolor}
\usepackage{subfigure}
\usepackage{spconf,amsmath,graphicx}
\usepackage{multirow}
\usepackage{hyperref}% For hyperlink
\usepackage{url}

\usepackage{enumitem}
 \usepackage{multirow}
\setlist{nosep, leftmargin=14pt}

\usepackage{mwe} % to get dummy images

\usepackage[skip=5pt]{caption}

\title{Towards Improved Cervical Cancer Screening:\\Vision Transformer-Based Classification and Interpretability}
\name{Khoa Tuan Nguyen$^{1,2}$, Ho-min Park$^{1,2}$, Gaeun Oh$^{1}$, Joris Vankerschaver$^{1,3}$, Wesley De Neve$^{1,2}$}

\address{$^1$ Center for Biosystems and Biotech Data Science, Department of Environmental Technology, \\Food Technology, and Molecular Biotechnology, Ghent University Global Campus, Incheon, Korea \\
$^2$ IDLab, Department of Electronics and Information Systems, Ghent University, Ghent, Belgium \\
$^3$ Department of Applied Mathematics, Computer Science and Statistics,Ghent University, Ghent, Belgium
}

\begin{document}
%\ninept
%
\maketitle

\begin{abstract}
We propose a novel approach to cervical cell image classification for cervical cancer screening using the EVA-02 transformer model.
We developed a four-step pipeline: fine-tuning EVA-02, feature extraction, selecting important features through multiple machine learning models, and training a new artificial neural network with optional loss weighting for improved generalization. With this design, our best model achieved an F1-score of 0.85227, outperforming the baseline EVA-02 model (0.84878). We also utilized Kernel SHAP analysis and identified key features correlating with cell morphology and staining characteristics, providing interpretable insights into the decision-making process of the fine-tuned model.
Our code is available at \url{https://github.com/Khoa-NT/isbi2025_ps3c}.
\end{abstract}

\section{Introduction}
Cervical cancer remains a significant global health challenge, ranking as the fourth most common cancer among women with over 600,000 new cases and 300,000 deaths annually~\cite{WHO_CervicalCancer}. 
Early detection through Pap smear screening has proven crucial in reducing mortality rates by identifying precancerous lesions.
However, traditional analysis methods face major challenges: they are resource-intensive, time-consuming, and heavily dependent on cytologist expertise. 
To address these challenges, we participated in the Pap Smear Cell Classification Challenge (PS3C)~\cite{kupas2024annotated,harangi2024ps3c}, organized as part of the IEEE International Symposium on Biomedical Imaging (ISBI) 2025 Challenge Program to foster the development of automated classification systems for cervical cell images.

In this challenge, we were tasked with developing deep learning models to classify Pap smear cell images into three categories: healthy cells without abnormalities, unhealthy cells indicating potential pathological changes, and unsuitable images due to artifacts or poor quality. 
The challenge provided a comprehensive training dataset containing four classes (including an additional `Both cells' category), while the test dataset focused on the three primary categories~\cite{kupas2024annotated,harangi2024ps3c}.
Effectiveness was evaluated using the F1-score, calculated for each class and averaged across all classes to account for data imbalance. 
% Our code and corresponding documentation are available at our GitHub repository\footnote{\url{https://github.com/Khoa-NT/isbi2025_ps3c}.}

\section{Methods}
\begin{figure*}[ht]
  \centering
\includegraphics[width=0.88\linewidth]{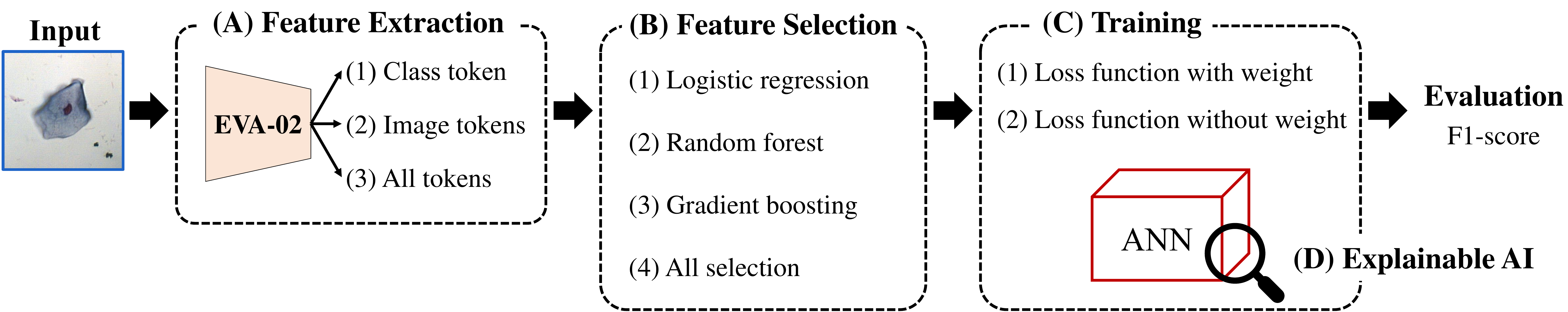}
  \caption{Overview of the experiment.}
  \label{fig:overview}
\end{figure*}

We used a four-step method to achieve high F1-scores.
First, we fine-tuned the pre-trained transformer-based image classification model EVA-02~\cite{eva02}. 
Then, as shown in Fig.~\ref{fig:overview}, we (A) extracted features from the model, (B) selected important features, and (C) trained a new Artificial Neural Network (ANN) model with the selected features with/without loss weighting for imbalanced labels. 
Finally, (D) we employed Kernel Shapley Additive Explanations (SHAP) to analyze the decision-making process of the fine-tuned model and provide interpretable insights into how the model makes its classifications~\cite{lundberg2017unified}.

\subsection{Fine-tuning EVA-02}

\begin{figure*}[htb]
    \centering
    \includegraphics[width=0.80\linewidth]{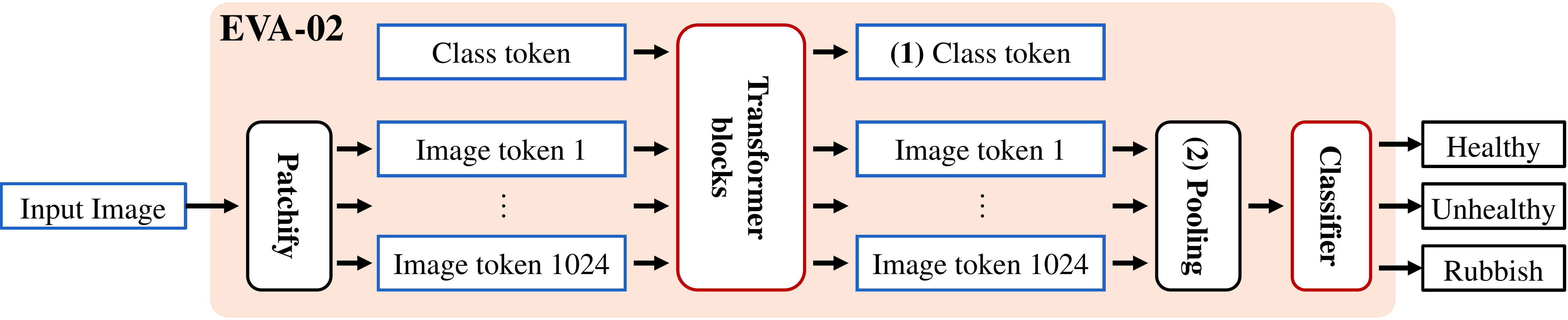}
    \caption{
    A simple illustrative workflow of the EVA-02 model.  
    The input cell image is divided into patches, creating Image tokens.  
    A sequence of transformer-based blocks processes a trainable Class token along with these Image tokens.  
    The output Image tokens are then average pooled before being fed into the classifier to predict Healthy, Unhealthy, and Rubbish classes.
    The Class token is extracted at (1), while the Image tokens are extracted at (2).
    }
    \label{fig:EVA-02}
\end{figure*}

We use EVA-02~\cite{eva02}, a transformer-based architecture, as our baseline classification model. 
The selection of EVA-02 was based on its superior performance among transformer-based image classification models available in PyTorch Image Models\footnote{According to the benchmark results: \url{https://github.com/huggingface/pytorch-image-models/blob/main/results/results-imagenet.csv}}~\cite{rw2019timm}.
Instead of training from scratch, we fine-tune the pre-trained ImageNet model\footnote{Model: `timm/eva02\_base\_patch14\_448.mim\_in22k\_ft\_in22k\_in1k'} on the entire training dataset to classify Healthy, Unhealthy, and Rubbish output classes, as shown in Fig.~\ref{fig:EVA-02}.
In addition, encouraged by the challenge organizers~\cite{harangi2024ps3c}, we found that training with three classes (merging the `Both cells' category into the `Unhealthy' category) produced slightly better results than training with four classes. 
Therefore, we adopt this merging as the default setting.
The same configuration as the pre-trained ImageNet model is used for image preprocessing and augmentation.
We employ the AdamW optimizer ($\text{weight decay}=0.05$) with a learning rate of $1\mathrm{e}{-5}$, a ReduceLROnPlateau scheduler ($\text{patience}=2$, $\text{factor}=0.5$), and CrossEntropyLoss from PyTorch~\cite{Ansel_PyTorch_2_Faster_2024}.
We fine-tune the model for 20 epochs and select the checkpoint with the lowest loss value.

% Extracting features
\subsection{(A) Feature extraction}

After completing the training, we extracted image features using three options: Class token, Image tokens, and All tokens. 
The Class token, shown as (1) in Fig.~\ref{fig:EVA-02}, is 
an additional token designed to capture comprehensive information from the images in the dataset~\cite{dosovitskiy2020vit}.
The Image tokens, which can be seen as (2) in Fig.~\ref{fig:EVA-02}, are generated by dividing the input image into patches, with each token representing local features containing both positional and visual information from its corresponding patch. 
The `All tokens' option includes both the Class token and Image tokens, providing the most comprehensive feature representation by combining both global and local information.

\subsection{(B) Feature selection}

% Feature masking method

% The features from EVA-02, namely Class token, Image tokens, and All tokens, each possess a dimensionality of 768. 
The features from EVA-02, namely Class token, Image tokens, and All tokens, share the same format shape $(N, L, E)$, where $N$ is the number of images in the training dataset, $L$ represents the token length, and $E$ denotes the embedding size. 
By averaging along the token length axis (with $L_{\text{Class token}}=1$, $L_{\text{Image tokens}}=1024$, and $L_{\text{All tokens}}=1025$), each feature extracted from an image obtains a universal dimensionality of $E=768$.
Given the limited amount of training data available relative to the image size, we trained these features with their corresponding labels (Rubbish, Healthy, Unhealthy) using three machine learning models (Logistic regression, Random forest, Gradient boosting) to reduce the risk of overfitting and eliminate less substantial features that could potentially introduce noise~\cite{Li2017}.

We extracted the importance of each feature from each trained model. 
For instance, in Logistic regression, the absolute value of coefficients served as the importance measure for the corresponding features. 
The importance values were averaged across the three classes to generate rankings, and feature selection was performed by applying different thresholds for each model:

\begin{itemize}
    \item Logistic regression: Determined through manual inspection, applied a manual threshold of $1\mathrm{e}{-16}$ (31.12\% filtered)
    \item Random forest: Applied a threshold of $3\mathrm{e}{-6}$, determined by manually identifying the point where the number of data points and their values decreased sharply in the distribution (1.95\% filtered)
    \item Gradient boosting: Excluded all data points with an importance value of 0 (40.10\% filtered)
\end{itemize}

\noindent This feature selection process enabled us to maintain model effectiveness while reducing the risk of overfitting and facilitating more efficient training. For comparison purposes, we also included an `All selection' option where all feature dimensions were used without any selection process.

\begin{table*}[!h]
    \centering
    \caption{
    F1-scores~$\uparrow$ (higher is better) for different combinations of feature extraction methods and machine learning models under weighted and unweighted loss conditions.
    The notable scores are shown in \textbf{bold}.
    }
    \resizebox{0.7\textwidth}{!}{
    \begin{tabular}{|c|c|c|c|}
        \hline
        (A) Extraction method & (B) Selection & (C-1) Weighted loss & (C-2) Unweighted loss \\ \hline
        \multirow{4}{*}{Class token} & Gradient boosting & 0.85007 & 0.85033 \\
        & Random forest & 0.85020 & 0.85112 \\
        & Logistic regression & 0.85031 & 0.84921 \\
        & All selection & 0.85039 & 0.85015 \\ \hline
        \multirow{4}{*}{Image tokens} & Gradient boosting & 0.85069 & 0.85037 \\
        & Random forest & 0.85070 & 0.85166 \\
        & Logistic regression & 0.85078 & 0.85141 \\
        & All selection & 0.85154 & 0.85108 \\ \hline
        \multirow{4}{*}{All tokens} & Gradient boosting & 0.85056 & 0.85044 \\
        & Random forest & \textbf{0.85225} & 0.85008 \\
        & Logistic regression & 0.85004 & 0.85072 \\
        & All selection & 0.85169 & \textbf{0.85227} \\ \hline
        \hline
        \multicolumn{2}{|c|}{EVA-02 baseline} & 0.84170 & \textbf{0.84878} \\ \hline
        
        \end{tabular}
    }
    \label{tbl:result}
\end{table*}

\subsection{(C) Training a new ANN model}

To integrate the extracted features into a classification model, we constructed an ANN with three hidden layers (1024, 512, and 256 neurons, respectively), choosing these values empirically based on their practical performance. 
Compared to the EVA-02 baseline classifier, which consists of a single linear layer (input to output:  $768 \rightarrow 3$), this deeper architecture allows for more expressive feature learning.

The ANN was trained for 100 epochs using cross-entropy loss, and we selected the checkpoint with the lowest validation loss for evaluation.

\textbf{Loss Weighting.}
Instead of treating all categories equally in the cross-entropy loss, we propose adding weights to the loss function for each category based on the number of data points, as shown in Table~\ref{tab:class_weight}.
We also fine-tune the baseline model with loss weighting for comparison.

The weighted cross-entropy loss function is defined as follows:

\[
\mathcal{L} = - \sum_{c=1}^{C} w_c \cdot y_c \cdot \log(p_c),
\]

where:
\begin{itemize}
    \item \( C \) is the number of classes.
    \item \( y_c \) is the binary indicator (0 or 1) for whether class label \( c \) is the correct classification for the current instance.
    \item \( p_c \) is the predicted probability of class \( c \).
    \item \( w_c \) is the weight for class \( c \), calculated as:
\end{itemize}

\[
w_c = \frac{\max(\text{num\_data\_points})}{\text{num\_data\_points}_c}
\]

where:
\begin{itemize}
    \item \( \max(\text{num\_data\_points}) \) is the maximum number of data points across all classes.
    \item \( \text{num\_data\_points}_c \) is the number of data points in class \( c \).
\end{itemize}

\begin{table}[h]
    \centering
    \resizebox{0.9\columnwidth}{!}{
    \begin{tabular}{|c|c|c|}
        \hline
        Class & \# of datapoints & Weights \\
        \hline
        Rubbish & 50,371 & 1.000 \\
        Healthy & 28,895 & 1.743 \\
        Unhealthy + Both cells & 5,814 & 8.664\\
        \hline
    \end{tabular}
    }
    \caption{Weight of each class.}
    \label{tab:class_weight}
\end{table}

\subsection{(D) SHAP-based feature analysis}
\label{sec:method:D}
While achieving high effectiveness is important, understanding model decisions is equally crucial, particularly in medical applications where explainability directly relates to patient trust and rights~\cite{goodman2017european}. 
We employed Kernel SHAP~\cite{lundberg2017unified} to analyze the decision-making process of our best model.
This analysis identifies the most influential feature among the input features and examines how variations in the values of this feature correlate with the actual image characteristics. 
We compared images where this important feature showed extremely low versus high values, providing qualitative insights into what visual patterns or characteristics this feature represents.

\section{Results and discussion}

\subsection{(A) Impact of feature extraction}
Our analysis indicates that the choice of feature extraction method had a notable impact on the effectiveness of the model. 
As shown in Table~\ref{tbl:result}, among the three extraction approaches (Class token, Image tokens, and All tokens), the All tokens method generally achieved the highest F1-scores, with a maximum effectiveness of 0.85227 obtained using 'All selection' under unweighted loss conditions. 
This suggests that the combination of both global (Class token) and local (Image tokens) information provides the most comprehensive representation for classification.

\subsection{(B) Feature selection effectiveness}
Our feature selection approach, as reflected in the results of Table~\ref{tbl:result}, which filtered out less significant features based on model-specific thresholds (40.10\% for Gradient boosting, 1.95\% for Random forest, and 31.12\% for Logistic regression), proved effective in maintaining model effectiveness while reducing computational complexity. 
The competitive effectiveness of models with filtered features suggests that our selection criteria successfully identified the most relevant features for classification.

\subsection{(C) Effect of loss weighting}
The impact of loss weighting on model effectiveness showed notable patterns in Table~\ref{tbl:result}. 
While the differences between weighted and unweighted loss scenarios were relatively small (differences typically less than 0.002), there were consistent patterns across different combinations of the extraction (A) and selection (B) methods. 
For instance, with Class token extraction, weighted loss generally led to more stable F1-scores across different machine learning models (range: 0.85007-0.85039) compared to unweighted loss (range: 0.84921-0.85112).

\subsection{Comparison with baseline EVA-02}
The results in Table~\ref{tbl:result} demonstrate that our proposed feature extraction and selection pipeline consistently outperformed the baseline EVA-02 model (F1-score: 0.84878). 
Our best performing configuration, using All tokens, All selection, and unweighted loss, achieved an F1-score of 0.85227, representing a 0.35\% improvement over the baseline.

\subsection{(D) SHAP-based feature analysis}

\begin{figure}[!ht]
    \centering
    \subfigure[Sample images with high feature \#10 value, characterized by large cells with rich cytoplasm and intense red/blue staining with clear nuclear boundaries.]{
        \centering
        \begin{minipage}{0.47\textwidth}
            \centering
            \includegraphics[width=0.45\textwidth]{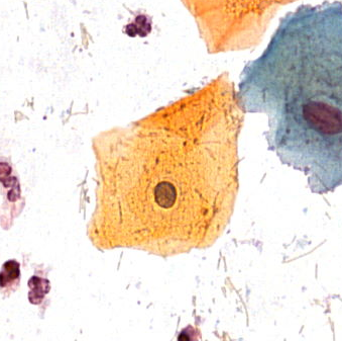}
            \includegraphics[width=0.45\textwidth]{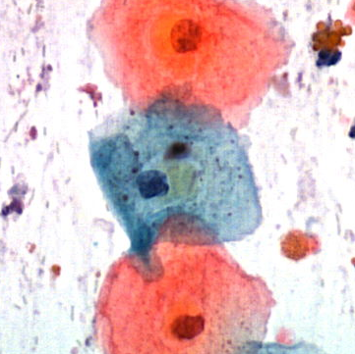}
            
        \end{minipage}
    }

    \subfigure[Sample images with low feature \#10 value, showing a large dark uncertainty region.]
    {
        \centering
        \begin{minipage}{0.47\textwidth}
            \centering
            \includegraphics[width=0.50\textwidth]{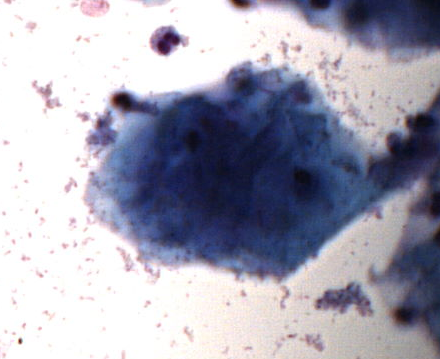}
            \includegraphics[width=0.42\textwidth]{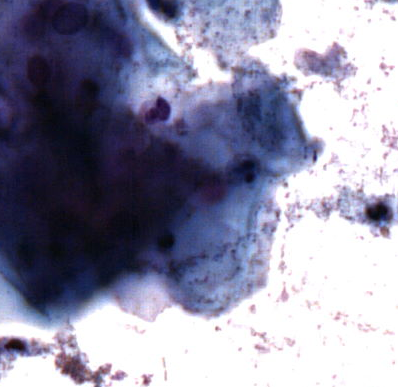}

        \end{minipage}
    }
    \caption{Comparison of cell images based on feature \#10 values, demonstrating distinct morphological and staining characteristics.}
    \label{fig:samples}
\end{figure}

We analyzed the relative importance of each feature among the 768-dimensional input features using our best model (All tokens + All selection + Unweighted loss).
This analysis demonstrated that feature \#10 emerged as the most influential feature in the decision-making process of the model.
%Notably, this feature also had the strongest positive correlation (0.77) with the Healthy class and negative correlations with other classes in the Logistic regression analysis from our feature selection process (B).% It turns out this description was based on the Class token, not the All tokens.. I removed it.

Fig.~\ref{fig:samples} provides visual insights into the characteristics of feature \#10. 
Fig.~\ref{fig:samples}(a) displays images with high feature \#10 values, while Fig.~\ref{fig:samples}(b) shows images with low values. 
The visual characteristics exhibit substantial differentiation: images with high feature \#10 values are characterized by large cells with rich cytoplasm and intense red/blue staining with clear nuclear boundaries, whereas images with low feature \#10 values show dark regions with indistinct cells.

\section{Conclusion}

In this study, we proposed an innovative approach for cervical cell image classification to aid in early cervical cancer detection. 
Our pipeline, which combines feature extraction based on the EVA-02 transformer model, feature selection, and a deeper ANN classifier, achieved higher F1-scores compared to the EVA-02 baseline.
A notable finding is that the All tokens approach, which utilizes both global information (Class token) and local information (Image tokens), demonstrated the highest effectiveness. Additionally, we confirmed that through the feature selection process, we could maintain effectiveness while reducing model complexity.
Furthermore, through Kernel SHAP, we identified feature \#10 as having the most impact on the decision-making of the model. Particularly, through image analysis, we were able to confirm that cell size, cytoplasmic characteristics, and staining intensity demonstrate notable influence in classification.

The main contribution of our research findings lies in presenting a method that satisfies both the high accuracy and interpretability required in medical settings for the automation of cervical cancer screening. However, a limitation of this study is that the difference between the minimum and maximum effectiveness is 1.06\% (minimum: 0.8417, maximum: 0.85227), and the final results are not yet confirmed as the evaluation has only been conducted on the public leaderboard.

%\section{Acknowledgments}
%This research was supported by Ghent University Global Campus (GUGC) in Korea and the Department of Electronics and Information Systems at UGent in Belgium.

\bibliographystyle{IEEEbib}
\bibliography{refs}

\end{document}